\documentclass{article}
\usepackage{ijcai22}

\usepackage{times}
\usepackage{soul}
\usepackage{url}
\usepackage{multirow}
\usepackage{makecell}
\usepackage[hidelinks]{hyperref}
\usepackage[utf8]{inputenc}
\usepackage[small]{caption}
\usepackage{graphicx}
\usepackage{amsmath}
\usepackage{amsthm}
\usepackage{booktabs}
\usepackage{algorithm}
\usepackage{algorithmic} 
\usepackage{array}
\usepackage{color}
\usepackage{amsfonts}
\usepackage{enumitem}
\title{Watermarking Pre-trained Encoders in Contrastive Learning}

\author{
Yutong Wu$^1$\footnote{Contact Author}\and
Han Qiu$^1$\and
Tianwei Zhang$^{2}$\and
Jiwei Li$^3$\and \And
Meikang Qiu $^4$
\affiliations
$^1$Tsinghua University, China\\
$^2$Nanyang Technological University, Singapore\\
$^3$Shannon.AI, Zhejiang University, China\\
$^4$Texas A\&M University, Texas, USA\\
\emails
wu-yt18@mails.tsinghua.edu.cn,
qiuhan@tsinghua.edu.cn,
tianwei.zhang@ntu.edu.sg,
jiwei\_li@shannonai.com,
qiumeikang@yahoo.com
}

\begin{document}

\maketitle

\begin{abstract}
Contrastive learning has become a popular technique to pre-train image encoders, which could be used to build various downstream classification models in an efficient way. This process requires a large amount of data and computation resources. Hence, the pre-trained encoders are an important intellectual property that needs to be carefully protected. It is challenging to migrate existing watermarking techniques from the classification tasks to the contrastive learning scenario, as the owner of the encoder lacks the knowledge of the downstream tasks which will be developed from the encoder in the future. We propose the \textit{first} watermarking methodology for the pre-trained encoders. We introduce a task-agnostic loss function to effectively embed into the encoder a backdoor as the watermark. This backdoor can still exist in any downstream models transferred from the encoder. Extensive evaluations over different contrastive learning algorithms, datasets, and downstream tasks indicate our watermarks exhibit high effectiveness and robustness against different adversarial operations. 

%Using contrastive learning, a model owner can train a powerful pre-trained model with his unique resource. This makes the model to be a intellectual property. To protect this intellectual property, watermarking was recently proposed. However, all existing works of watermark focus on end-to-end classification tasks to the best of our knowledge, and thus difficult to deal with task-agnostic scenarios. In this work, we propose a new method to embed watermark into a pre-trained model without notarizing its downstream task and get a watermarked encoder as consequence. We regard our watermarked encoder as a optimization problem and use gradient decent to solve it. The result of our experiments on several data sets tested as downstream tasks show that our watermark is of high effectiveness. We also test some considerable attacks such as fine-tuning. Our result shows that our method is rather robust toward those attacks. Finally, the uniqueness of our watermark is proved to be a first-class one by the experiment, indicating the possible practicability of our method.
\end{abstract}

\section{Introduction}

Contrastive learning \cite{chen2020simple,chen2020improved,caron2020unsupervised,chen2020exploring,grill2020bootstrap,zbontar2021barlow} has demonstrated huge potential for unsupervised learning recently. 
Its main idea is to use a large number of unlabeled data to pre-train an encoder for effective feature representation, which could be further used to build different types of downstream classifiers with limited labeled data. 
Such approaches can be performed over noisy and uncurated data which can get rid of the expensive data labeling efforts~\cite{carlini2021poisoning}. 
Moreover, contrastive learning has achieved better performance than classic supervised learning in many tasks. 
For instance, SimCLR \cite{chen2020simple}, a popular contrastive learning algorithm, has been proved to have a better performance in an ImageNet classification task than various supervised learning algorithms. 

A typical contrastive learning pipeline consists of two components, i.e. pre-training encoders and building downstream classifiers. 
As the most important component, the pre-trained encoders are usually generated with many unlabeled data and a large number of training costs (e.g., GPU resources). 
Moreover, for special tasks like medical diagnose, collecting critical training data is also a costly process. 
As a result, well-trained encoders are very valuable since they can be applied to many downstream classification tasks with a simple fine-tuning process and a small number of labeled data. Selling pre-trained encoders has become a popular business model \footnote{\url{https://twimlai.com/solutions/features/model-marketplace/}}, and these encoders are an important intellectual property (IP) for model owners. It is necessary to protect these models from being abused, illegal copy, and redistribution.

One promising technology to protect the IP of a DNN model is watermarking~\cite{adi2018turning,zhang2018protecting,fan2019rethinking,zhang2020model}. A quantity of works have designed watermarking solutions for conventional classification models. 
By embedding a backdoor into the protected classification model during the training process, the model owner can easily verify the ownership of any suspicious model via remote query using the specific triggered samples. 
%Using an embedded backdoor into one DL classifier during the training process, the IP of this classifier can be easily verified by querying with a triggered sample. 
However, it is challenging to directly apply a similar approach to watermark a pre-trained encoder from contrastive learning due to two reasons. 
First, when embedding the watermarks, the owner of the pre-trained encoder does not know the specific downstream classification tasks that will be built from it, as well as the corresponding dataset. Therefore, it is difficult for the owner to craft backdoor and verification samples. 
%However, during verification, he can only query each downstream model to check if the watermark exists in it. Such a gap makes it difficult to implant watermarks using prior techniques. 
%It is known how to implant watermarks unified for arbitrary downstream tasks and datasets. 
Second, during verification, the encoder owner can only query the downstream model to check if it is built from his watermarked encoder. He can only obtain the final output from the classification layer, rather than the feature representation from his encoder. This also hinders the owner from checking the existence of a backdoor.

%he can only query each downstream model to check if the watermark exists in it. when verification is based on the watermark, the verifier does know the output of the tested encoder since the downstream classifier has been built. The verification is impossible by only knowing the final prediction results of the downstream classifiers which are after through some extra layers of the pre-trained encoder. 

%\noindent\textbf{Our solution:} 

To address these challenges, this work proposes the first watermarking methodology for the IP protection of the pre-trained encoders in contrastive learning. Our solution can achieve \textit{task-agnostic}, i.e., the owner does not need any prior information about the downstream models and datasets, and the embedded watermark is verifiable for arbitrary downstream tasks.  
Particularly, our watermark is also based on the backdoor technique. Instead of manipulating the verification samples to have unique predicted labels, we aim to make them have unique feature representations from the encoder, which can naturally lead to unique predicted labels for any subsequent downstream models. Specifically, we introduce a loss function to fine-tune the model for watermark embedding, which can make its output of samples with a specially-designed trigger deviate a lot from the output of a normal encoder. When the adversary illegally obtains this encoder and trains a classification model from it, the owner can use verification samples (i.e., normal samples with the trigger) to query the model. When its label is different from the normal case (i.e., query results of the same normal sample), the owner has confidence to identify this as a plagiarization.

We extensively evaluate our watermarking method on different popular datasets (STL10, GTSRB, and SVHN) with two representative contrastive learning algorithms (SimCLR \cite{chen2020simple} and MocoV2\cite{chen2020improved}). 
The results indicate that our proposed technique is effective at distinguishing plagiarized models from independent ones regardless of the downstream tasks. It will not affect the functionality of the encoder and any downstream models. Besides, the embedded watermark exhibits strong robustness against different adversarial operations (e.g., fine-tuning, pruning), making it hard to be removed.

%is effective on the both algorithms and \textit{preserves functionality} for bengin samples. It also shows that our watermark is of a high \textit{uniqueness} that the watermarked model can be verified successfully on all of the downstream datasets {\it only} by the corresponding verification samples. Finally, it turns out that our watermark is \textit{robust} considering potential attacks like fine-tuning and pruning.

\section{Background}

\subsection{Contrastive Learning}
The aim of contrastive learning is to 
pre-train an encoder by using a huge amount of unlabeled images. Specifically,
for each image $\varphi$, the contrastive learning algorithm first generates a positive sample $\varphi^{+}$ which is transformed from $\varphi$, as well as a negative sample $\varphi^{-}$, which is transformed from a different sample. 
We expect the outputs $f(\varphi)$ and $f(\varphi^{+})$ to be as similar as possible, since $\varphi$ and $\varphi^{+}$ are from the same class.
In addition, we also need to maximize the difference between $f(\varphi)$ and $f(\varphi^{-})$, as they are likely to be in different classes. 
By pursuing the above optimization objective, the encoder can be trained with unlabeled samples to accurately predict the feature representation of any image from any category. Below we briefly introduce two representative contrastive learning algorithms. 

\vspace{3pt}
\noindent\textbf{SimCLR}~\cite{chen2020simple}. This algorithm uses common data augmentation operations (random crop, Gaussian blur, random flip, etc.) to generate positive and negative samples. 
For a $N$-image batch, SimCLR generates $2N$ images $(x_i, (i=1,...,2N))$ by applying data augmentation twice to the samples in it. 
Two transformed images form a positive pair if they are originated from the same image, or a negative one otherwise. 
To achieve the goal of contrastive learning, a contrastive loss is defined as below:
\begin{equation}
\label{eq: contrastive loss of SimCLR1}
l_{i,j} = -\log\Big(\frac{\exp({sim(z_i,z_j)/\tau})}{\sum^{2N}_{k=1}{\mathbb{I}(k\neq i)\cdot \exp({sim(z_i, z_k)/\tau})}}\Big)
\end{equation}
where $(z_i,z_j)$ is the positive pair of feature vectors produced by the encoder (here $z_i = f(x_i)$), while $(z_i,z_k)$ is the negative one. 
$sim(\cdot, \cdot)$ is the cosine similarity between the two feature vectors. 
$\tau$ is a temperature parameter used to scale the cosine similarity and $\exp$ is the natural exponential function. 
If the feature vectors of the positive pair ($z_i, z_j$) are more similar and the ones of the rest negative pairs ($z_i, z_k$) are more different, the loss value will be smaller. 
Therefore, we need to update the encoder parameters to minimize the above loss function. 
SimCLR sums up the loss of all positive pairs from the batch as the final loss and learns the image encoder via minimizing this final loss. 

\vspace{3pt}
\noindent\textbf{MocoV2}~\cite{chen2020improved}. The core of this algorithm is to introduce a dictionary of feature vectors as a {\it queue} so as to reduce the memory cost in the training process by reusing the feature vectors from the immediate mini-batches. 
To this end, MocoV2 uses two encoders called query encoder ($f_q$) and momentum encoder ($f_k$). 

Given a batch of $N$ input images, $f_q$ produces $N$ feature vectors $q$ and $f_k$ produces $N$ feature vectors $k_+$ by applying different data augmentations on the input images. 
The contrastive loss is then subsequently calculate as below: 
\begin{equation}
\label{eq: contrastive loss of SimCLR}
\mathcal{L} = -\log\Big(\frac{\exp({q\cdot k_+/\tau})}{\sum^{K}_{i=0}{\exp(q\cdot k_i/\tau)}}\Big)
\end{equation}
where $k_i$ is the feature vectors of previous batches. 
$\tau$ is the same as in SimCLR. 
Usually, the queue has a much bigger size than the mini-batch. 
After calculating the contrastive loss, the feature vectors produced by the momentum encoder ($k$) are then pushed into the queue and one of the earliest batches is deleted simultaneously.

MocoV2 updates the parameters of its two encoders in different ways. 
The parameters in the query encoder are updated by the back-propagation algorithm according to the immediate contrastive loss. The parameters in the momentum encoder are updated by the following Equation:
\begin{equation}
\label{eq: parameter updating algorithm of the momentum encoder}
\theta_k = m\cdot\theta_k' + (1-m) \cdot \theta_q 
\end{equation}
where $\theta_k'$ is the old parameters of the momentum encoder, $\theta_q$ is the immediate parameters of the query encoder, $m$ is a hyper-parameter to control the updating speed, and $\theta_k$ is the updated parameters of the momentum encoder.

\subsection{Watermarking}
\textbf{Watermarking DL Models.}
The general goal of watermarking is to protect the IP of a DNN model. The most popular watermarking solution leverages the DNN backdoor technique. 
A standard watermarking solution consists of two phases~\cite{guo2021fine}. 
In the first phase, the model owner employs a watermark embedding algorithm to inject a backdoor into the target DL model to get a watermarked classifier. 
This watermarked classifier maintains the functionality for normal samples while giving unique labels for some carefully-crafted samples (i.e., verification samples).
In the second phase, the model owner attempts to verify whether a suspicious classifier contains his watermark. He uses the verification samples to query this classifier. By checking the corresponding responses, he can identify the ownership of this model with high confidence.
%The second component is a verification scheme. Normally, a triggered sample corresponding to the watermark embedding algorithm will be fed into the suspicious classifier to verify the IP by checking the inference result. 
%A successful watermarking method 

%By doing that, we get a watermarked copy of the original data $D_w=D\oplus W$. The watermarked copy might be slightly different to its original, but it maintains utility. The other component is called verification. The verification algorithm extract watermarks in suspicious data $D'$ and get its watermark $W'$. By comparing $W'$ and $W$, the algorithm can distinguish whether the suspicious data are the copy of the original ones.

\vspace{3pt}
\noindent\textbf{Threat model.} 
%Our threat model is consisted of two parts. 
We consider a model owner $M$ who pre-trains an encoder $f$ and sells it to some users for building downstream tasks. However, a model plagiarist $P$ gets this encoder in an unauthorized way (e.g., model stealing, illegal redistribution, etc.). 
%has the right to authorize downstream users to use $f$ without any knowledge about the downstream tasks. A model plagiarist $P$ who may get $M$'s encoder in an unauthorized way such as model stealing attack of bargaining on the dark net. 
How the adversary obtains the encoder is beyond the scope of this paper. 
%Note that how the pre-trained encoder leaks or being stolen is not within the scope of this paper. 

Then the plagiarist uses $f$ to build his own downstream classifier. He may slightly modify the encoder (e.g., fine-tuning, pruning). We assume the adversary does not have enough resources to alter the encoder completely. Otherwise, he will train the encoder directly without stealing it. 

The owner's goal is to detect whether a suspicious classifier is built from his encoder. The model owner has only oracle access to this classifier, i.e., he can send arbitrary inputs to $F^s$ and receive the corresponding outputs. His strategy is to embed a watermark into $f$. When the classifier is from $f$, the classification output will be different from the ones output by an independent model. When embedding the watermark, the owner does not have any knowledge about the possible downstream tasks and datasets. The embedded watermark must be robust enough and unremovable by the possible model transformations from the adversary.

%We try to help the model owner to protect IP of $f$ by using watermark $W$ in this paper. Assuming $M$ finds a suspicious model $f_s$ which has a similar performance as his model $f$, he can use the verification algorithm $V$ to check whether it's from one of his authorized downstream users. 
%If the answer turns out to be yes, he can claim the proprietary right of the model. However, it is hard for $M$ to acquire full access to $f_s$. In our assumption, $M$ can only get the prediction results of $f_s$ and choose the input $I$.

%To realize the above IP verification, we propose our watermark-embedding algorithm to help $M$ to embed a watermark by using data only from the pre-training dataset and verify the watermark by only checking the prediction results of $f_s$. 
%Our watermarked model will produce an anomalous prediction result $O_a$ if the input $I$ is embedded with a special trigger $t$, which will lead to a mis-classification to $I$. 
%As $M$ knows the correct label of $I$, he can ascertain that $f_s$ originates of his pre-trained encoder from this mis-classification.

\vspace{3pt}
\noindent\textbf{Watermark Requirements.}
Specifically, a good watermarking solution should satisfy the following properties:
An effective watermark should meet several requirements to guarantee its usability. In our scenario, the watermark must have characters narrated as below: 
\begin{itemize}[leftmargin=*, itemsep=0pt, topsep=0pt]
    \item \textbf{Uniqueness.}
    This means the classification model from the watermarked encoder will give unique output for the verification samples different from that of an independent model. This is the basic requirement to guarantee the effectiveness of the watermarking scheme. 
%    This means a watermarked model can only be verified by the corresponding triggered sample and will not be able to be verified by triggered samples corresponding to other watermarked models. 
%    Also, a clean model without nay watermarks should not be verified by any triggers at all. 
    \item \textbf{Functionality-preserving.} 
    The watermarking algorithm should have a negligible impact on the performance of the downstream model to preserve its functionality.
    \item \textbf{Robustness.} 
    A successful watermark should be robust against potential attacks from the plagiarist, e.g., model modification via fine-tuning and pruning. Once embedded, the watermark is hard to be removed by the plagiarist. 
\end{itemize}

\begin{figure*}
    \centering
    \includegraphics[width=0.9\textwidth]{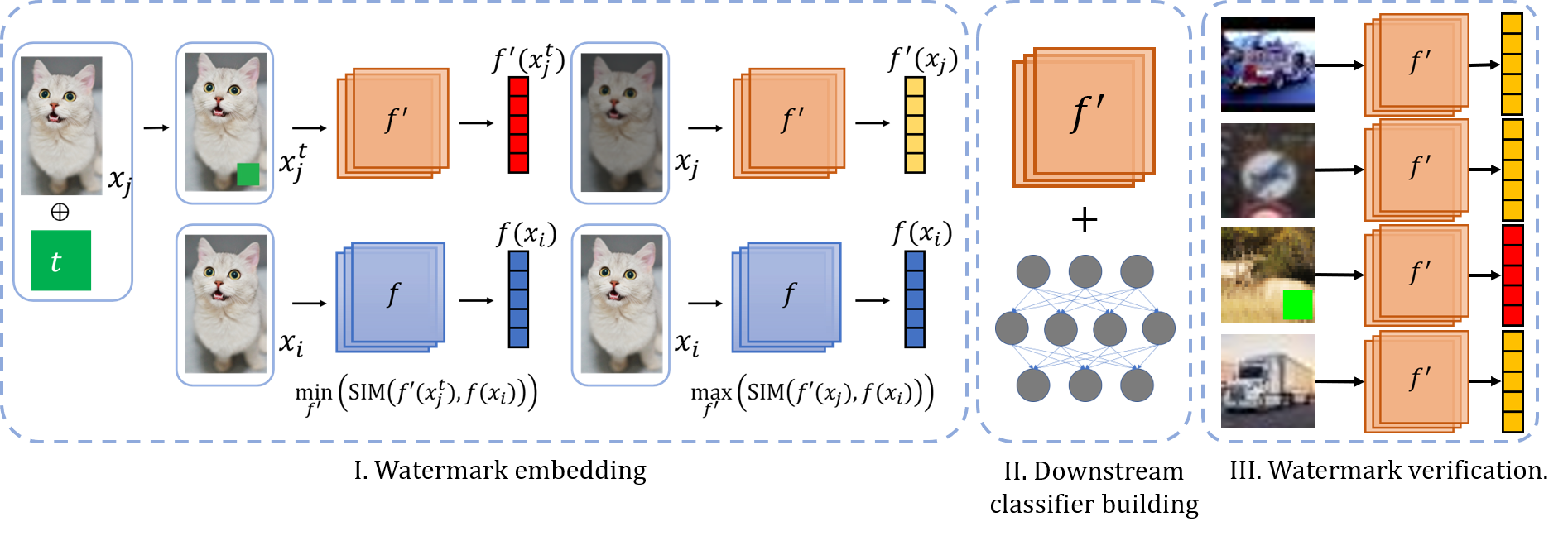}
    \vspace{-2ex}
    \caption{Overview of our watermarking technique}
    \label{fig:1}
\end{figure*}

\section{Methodology}

\subsection{Overview}

The most popular way to watermark a DL model is the DNN backdoor technique \cite{adi2018turning,zhang2018protecting}. In this paper, we also follow this idea for protecting the pre-trained encoders. Prior works have introduced backdoor attacks against pre-trained feature encoders \cite{jia2021badencoder} and language foundation models \cite{zhang2021trojaning}. However, these attacks require the adversary to have knowledge of downstream tasks and training sets. Therefore, they are not applicable to our scenario, where the encoder owner does not know anything about the downstream tasks and aims to protect the IP of any models developed from his pre-trained encoder. 

To this end, we design a new task-agnostic watermarking solution for pre-trained encoders. Figure \ref{fig:1} shows an overview of our methodology. The key insight is that instead of embedding a \textit{targeted} backdoor into the encoder that forces its downstream model to assign a specific label to the triggered samples, we can consider an \textit{untargeted} backdoor, causing the downstream model to output any labels that are different from the correct ones for the triggered samples. In this case, the encoder owner is still able to verify the ownership by checking whether the model output of the triggered samples is incorrect. Meanwhile, he can achieve such untargeted backdoor embedding without any knowledge of the downstream tasks. Below we detail our methodology. 

%Our aim is to watermark an clean encoder to achieve 1) uniqueness, and 2) robustness goal. To achieve the uniqueness goal, the watermarked encoder should be 1) sensitive to a given watermark embedded in an arbitrary picture, and 2) ignorant to any other watermarks. Consequently, a downstream classifier built based on our watermarked encoder gives anomalous outputs when inputs are embedded with a 'right' trigger, while preserves its utility with ordinary inputs and 'wrongly' embedded inputs. To achieve the robustness goal, a watermarked encoder should maintain its sensitivity towards a given trigger after being fine-tuned or pruned. Therefore, a fine-tuned downstream classifier based on either a pruned watermarked encoder or a simply watermarked one should keep its accuracy as well as its uniqueness to give ordinary outputs with clean pictures and wrong outputs with 'right' triggers.

\subsection{Watermark Embedding}
Formally, we consider a pre-trained encoder $f$, which takes as input an image, and outputs its corresponding feature. The owner aims to convert $f$ into a watermarked version $f'$, which can satisfy the properties of uniqueness, functionality-preserving, and robustness. The owner has access to the unlabeled training set of the encoder: $\mathcal{D}=\{x_{1}, x_{2},...,x_{n})\}$. 

To embed the watermark into $f$, the owner pre-defines a trigger pattern $t$ and a trigger mask $m$ to denote its position. Then for each sample $x_i \in \mathcal{D}$, he can calculate the corresponding triggered sample as in Eq.~\ref{eq:triggered sample}.
\begin{equation}
\label{eq:triggered sample}
x^t_i = (1-m)\otimes x_i + m \otimes t
\end{equation}

We craft the following loss function for the owner to fine-tune $f$ for watermark embedding. 

%To get an embedded encoder, firstly we should get a pre-trained one denoted as $f$. We call it {\it the clean encoder}. In our scenario, the clean encoders are pre-trained by a model owner and will be used in some downstream tasks after watermarking. {\it The watermarked encoders} which are denoted as $f'$ are what the downstream users are provided. Given that the model owner is unable to ascertain the downstream task, the only available data set is the one used to train the clean encoder.

%We denote the shadow data set as  $\mathcal{D}=\{x_{1}, x_{2},...,x_{n})\}$, where $x_{i}\space(i=1,...,n)$ is samples from the shadow data set, $n$ is the number of the samples. $w_t$ is the watermark that the owner chooses for task $t$. Operation $x\oplus w_i$ represents to embed the watermark $w_i$ in the chosen picture x. To better craft our watermarked encoder, we propose a two-term loss function. By doing that, we formulate the uniqueness goal and the robustness goal into an optimization problem.

\vspace{3pt}
\noindent\textbf{Uniqueness.}
To guarantee uniqueness, we need to ensure the downstream models developed from a normal encoder (e.g., $f$) and watermarked encoder ($f'$) have distinct predictions for each triggered sample $x^t_i$. As the owner does not know the downstream tasks and datasets, he cannot design loss functions to restrict the downstream model predictions, as did in \cite{jia2021badencoder}. Instead, he can try to maximize the difference of the encoders' output (i.e., features), which could subsequently maximize the difference of the downstream model's output (i.e., labels) with very high probability. This leads to the loss term in Eq.~\ref{eq:loss-uniqueness}.
\begin{equation}
\label{eq:loss-uniqueness}
\mathcal{L}_{u}=\frac{1}{||\mathcal{D}||}\cdot \sum_{x_i \in \mathcal{D}}{\texttt{SIM}\big(f(x_i), f'(x^t_i)\big)}
\end{equation}
where $\texttt{SIM}(\cdot,\cdot)$ measures the similarity between two feature vectors. In this paper, we adopt the cosine similarity in this loss term. Other similarity metrics can be applied as well. Our goal is to minimize $\mathcal{L}_u$, which can increase the feature distance between normal and watermarked encoders over triggered samples. 

%by calculating the cosine coefficient of their contained angle, $||\mathcal{D}||$ represents the number of samples in the shadow data-set. $L_1$ gets bigger if the watermarked encoder gives more similar feature vectors as outputs and smaller if the feature vectors differ. In the scenario of ours, we try to minimize it so as to embed our watermark into the encoder.

%\textbf{Uniqueness goal}: To achieve the uniqueness goal, the clean encoders must be invulnerable to the watermarks that we use. If the watermarks only have an imperceptible impact on the clean encoder, it will probably do the same in other-watermarked encoders. Therefore, watermarks of simple patterns are used. In other words, the watermarks we choose will only influence the outputs of the encoders slightly. The other concern of the uniqueness goal is its effectiveness to the watermarked encoder. Thus, we propose the $L_1$ term:

%$$L_1=\frac{1}{||\mathcal{D}||}\cdot \sum_{i=1}^{||\mathcal{D}||}{SIM\big(f(x_i), f'(x_i \oplus e_t)\big)}$$
%where $SIM(\cdot,\cdot)$ measures the similarity of two feature vectors by calculating the cosine coefficient of their contained angle, $||\mathcal{D}||$ represents the amount of samples in the shadow data-set. $L_1$ gets bigger if the watermarked encoder gives more similar feature vectors as outputs and smaller if the feature vectors  differ. In the scenario of ours, we try to minimize it so as to embed our watermark into the encoder.

\vspace{3pt}
\noindent\textbf{Functionality-preserving.}
We also need to guarantee the embedded watermark does not affect the prediction accuracy of any inherited downstream models over clean samples. Similarly, since the encoder owner does not know the downstream tasks, he can try to make the output feature of $f'$ as close as that of the normal encoder $f$ for clean samples, which will result in comparable performance in the subsequent downstream tasks. This gives us another loss term for functionality-preserving in Eq.~\ref{eq:loss-functionality}.
\begin{equation}
\label{eq:loss-functionality}
\mathcal{L}_p = -\frac{1}{||\mathcal{D}||}\cdot \sum_{x_i \in \mathcal{D}}{\texttt{SIM}\big(f(x_i), f'(x_i)\big)}
\end{equation}

%\textbf{Robustness goal}: To achieve the robustness goal, our watermarked encoder should preserve its uniqueness towards its watermark. One of the threats against this goal is pruning. We use random dropout in our training process to produce our watermarked encoder. By doing so, we simulate random pruning while training so the watermark may not be too vulnerable to it. The other threat is fine-tuning. The watermarked encoder may lose its efficacy to recognize the trigger. We propose $L_2$ term to solve the problem: $$L_2 = -1\cdot\frac{1}{||\mathcal{D}||}\cdot \sum_{i=1}^{||\mathcal{D}||}{SIM\big(f(x_i), f'(x_i)\big)}$$ Here $L_2$ is smaller if the watermarked encoder produces more similar feature vectors. The purpose of setting $L_2$ term is 1) to preserve the utility of the watermarked encoder and 2) to keep the similarity of the parameters between the watermarked encoder and the clean one. If the parameters of the watermarked encoder are similar to the clean one, it means that parameters used to embed watermarks may be less important to feature extracting. Therefore, they will not be modified very much in the fine-tuning process.

Based on the above analysis, the watermark embedding process can be formulated as an optimization problem as shown in Eq.~\ref{eq:loss-final}.
\begin{equation}
\label{eq:loss-final}
\min_{f'}({\mathcal{L}_u + \eta\cdot \mathcal{L}_p})
\end{equation}
where $\eta$ is the hyper-parameter to balance the properties of uniqueness and functionality-preserving. In this paper, we adopt the gradient descent method to solve the above optimization problem. We initialize $f'$ as the clean encoder $f$, and iteratively calculate the loss and update $f'$ to reach the optimal values.

%\textbf{Our optimization problems}: After the discussion and definitions above, we can regard our watermarked as the solution of the following optimization problem: $$\min_{f'}L={\eta_1\cdot L_1 + \eta_2\cdot L_2}$$ where $\eta_1$, $\eta_2$ are the hyperparameters to control the balance of the two loss function terms. A bigger $\eta_1$ means the encoder will be more sophisticated to recognize the corresponding trigger. Whereas a bigger $\eta_2$ will lead to a more robust encoder.

%\subsection{C. Solving the Optimization Problem}

%In our work, we use gradient descent to solve the optimization problem. The watermarked encoder is initiated as the clean one. We subsequently calculate the loss $L$ and upgrade the encoder slightly. The encoders originated from the two contrast learning algorithms are trained using different training epochs and batch sizes. 

\vspace{3pt}
\noindent\textbf{Robustness.}
As backdoor attacks generally exhibit high robustness against common model transformations (e.g., model pruning, fine-tuning), the corresponding watermarks are expected to enjoy similar robustness as well. We validate this conclusion in Section \ref{sec:evaluation}. In order to further enhance the watermark robustness in the encoder, we propose to adopt dropout during the embedding process, e.g., randomly dropping some neurons in each layer. This can simulate the impact of downstream model transfer learning and post-processing, which makes our watermarks more immune to these operations. 

\subsection{Watermark Verification}
Given a suspicious downstream model $M$, the encoder owner wants to verify whether $M$ is developed from $f'$ without IP authorization. To achieve this, he first constructs a set of data samples $\hat{\mathcal{D}}$ corresponding to this downstream task (e.g., via downloading from the internet). For each sample $\hat{x}_i \in \hat{\mathcal{D}}$, he feeds it to $M$ and obtains the predicted label. Then he calculates the corresponding verification sample $\hat{x}^t_i$ following Equation \ref{eq:triggered sample} and uses it to query $M$ again. He checks whether the two predictions are different. If the ratio of samples getting different labels in the two cases is higher than a pre-defined threshold $\mathcal{T}$, then the owner has confidence to confirm model $M$ violates the IP of his encoder $f'$. This process is formulated as in Eq.~\ref{eqn:verification}.
\begin{equation}
\label{eqn:verification}
\frac{1}{||\hat{\mathcal{D}}||} \cdot \sum_{\hat{x_i}\in \hat{\mathcal{D}}} \mathbb{I}(M(\hat{x_i}) \neq M(\hat{x_i}^t))) > \mathcal{T}
\end{equation}
where function $\mathbb{I}$ returns 1 when the inside condition is true, or 0 otherwise.

\section{Evaluation}
\label{sec:evaluation}
We perform extensive evaluations to demonstrate our solution can achieve the desired watermark requirements.

\noindent\textbf{Contrastive learning methods}.
Our solution is general for the encoders pre-trained from different contrastive learning methods. 
In this paper, we adopt SimCLR \cite{chen2020simple} and MocoV2 \cite{chen2020improved}, two of the most popular contrastive learning techniques for experimentation. 
We choose ResNet18 and ResNet50 as the base models for SimCLR and MocoV2, respectively.

\subsection{Experimental Setup}

%\vspace{3pt}
\noindent\textbf{Datasets}. We use different types of datasets to pre-train the encoders and build the downstream classifiers:
\begin{itemize}[leftmargin=*, itemsep=0pt, topsep=0pt]
    \item \textbf{CIFAR-10}: it contains 60,000 images belonging to 10 classes. 
    Each image has a size of $32\times32\times3$. We use 50,000 samples for training and 10,000 samples for testing. 

    \item \textbf{ImageNet}: it contains 14,000,000 images belonging to 1,000 classes. Each picture has a size of $224\times224\times3$. 
    %In our experiment, we don't use this data set directly, but we do use the pre-trained model provided by the author of MocoV2, which is trained using Imagenet.
    \item \textbf{STL10}: this dataset contains 113,000 images with the size of $96\times96\times3$. It has 5,000 training images and 8,000 test images with labels (10 classes). Besides, it also has 100,000 unlabeled images for unsupervised learning.
    \item \textbf{GTSRB}: This dataset contains 51,800 images of 43 different traffic signs. Each image has a size of $32\times32\times 3$.
    \item \textbf{SVHN}: it contains over 600,000 images of the digit house numbers from Google Street View. Each image has a size of $32\times32\times3$. It is divided into a training set of 73,257 images, a test set of 26,032 images, and an extra training set of 531,131 images. All the images belong to 10 classes corresponding to the digits from `0' to `9'.
\end{itemize}

In our experiment, we adopt CIFAR-10 and ImageNet for the pre-trained SimCLR and MocoV2\footnote{Instead of training the MocoV2 encoder from scratch, we use the model released by the authors directly at \url{https://github.com/facebookresearch/moco}} encoders, respectively. Then, we transfer the encoders to different downstream tasks for recognizing STL10, GTSRB, and SVHN samples. 

\begin{figure}[htbp]
    \centering
    \includegraphics[scale=0.5]{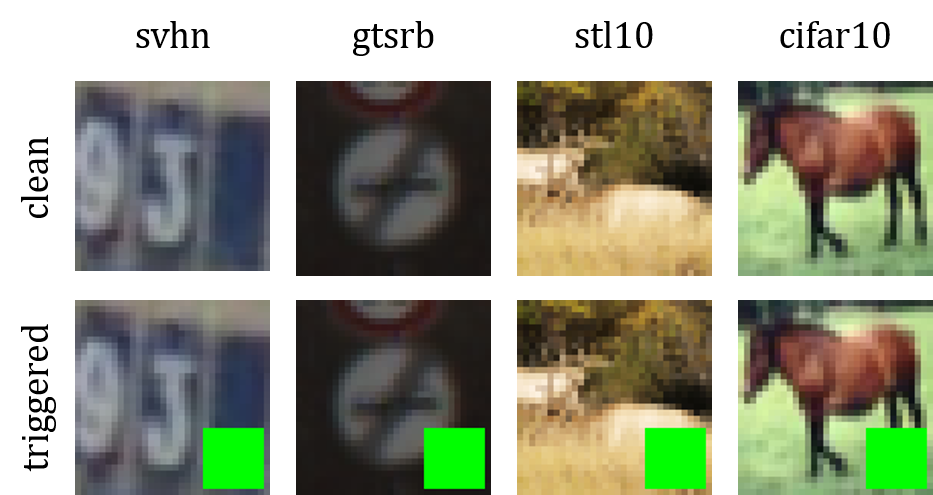}
    \caption{Examples of clean and triggered samples.}
    \label{fig:example}
\end{figure}

\noindent\textbf{Backdoor triggers.}
Our watermark method has no assumption on the triggers plant on samples because it does not rely on a special trigger to verify. 
Hence, in our experiments, we adopt the checkerboard trigger of size $10\times 10$ located on the bottom-right corner of images, which is similar to the Badencoder~\cite{gu2019badnets}. Specifically, We use white square, green square, and green cross as the triggers in our experiment.
Examples are in Figure~\ref{fig:example}: the first row represents clean samples and the second row contains triggered ones. 
Note that similar results will be get for different triggers.

%\vspace{3pt}
\noindent\textbf{Implementation.}
When embedding watermarks into the encoders, we adopt a batch size of 128 for SimCLR and train the encoder for 300 epochs. 
For MocoV2, we use a batch size of 64 and train the encoder for 50 epochs. 
For both training processes, we set the learning rate as 0.001 and $\eta$ as 0.5. 
We use Pytorch 1.10 backend for the implementation. 
We conduct the experiments on a server equipped with Intel I9-11900K CPU and 2 NVIDIA GeForce RTX 3090 GPUs. 

%The encoders originated from the two contrast learning algorithms are trained using different training epochs and batch sizes. We use a 128 batch size and train for 300 epochs for SimCLR encoders, while a 64 batch size and 50 epochs for MocoV2 encoders. In both cases, we set the learning rate to be 0.001 and $\eta_1/\eta_2$ to be 2.

\noindent\textbf{Possible attacks}. We consider two techniques including fine-tuning and pruning that the attacker may deploy to change the models for erasing the watermark. 
They are commonly adopted in the previous evaluation of watermark robustness \cite{adi2018turning,zhang2018protecting,guo2021fine}. 
\begin{itemize}[leftmargin=*, itemsep=0pt, topsep=0pt]
    \item \textbf{Fine-tuning}. We consider two types of fine-tuning methods in our experiments: {\it Re-Train Last Layers} (RTLL) and {\it Fine-Tune All Layers} (FTAL). 
    Note that we do not consider the operation of {\it Fine-Tune Last Layer} (FTLL) since the encoder only produces the feature vectors and the users always re-train the last layers for building downstream classifiers. 
    We also do not consider {\it Re-Train All Layers} (RTAL), which makes the pre-training meaningless. 
    \item \textbf{Pruning}. We evaluate two pruning methods: random pruning which randomly removes some parameters in each layer and L1-pruning which removes the parameters with the smallest L1-norms.
\end{itemize}

%\vspace{3pt}
\noindent\textbf{Metrics}. We adopt the following two metrics to evaluate the effectiveness of our watermark. 
\begin{itemize}[leftmargin=*, itemsep=0pt, topsep=0pt]
    \item \textbf{Test Accuracy (ACC)}. 
    This is to calculate the prediction accuracy of the downstream classifiers over clean samples. 
    This metric is used to measure the functionality-preserving requirement of the watermark: a downstream classifier built from a watermarked encoder should have a tiny ACC loss compared to the one built from a clean encoder. 
    
    \item \textbf{Watermark Accuracy (WACC)}. 
    This is to measure the ratio of verification samples' prediction labels by the suspicious model different from the prediction labels of the corresponding clean samples (Eq.~\ref{eqn:verification}). 
    This metric can evaluate the uniqueness requirement: the WACC of a watermarked downstream classifier should be much higher than an independent model which can be easily separated by a threshold $\mathcal{T}$. 
    Besides, WACC can also reflect the robustness requirement. 
    Under different attacks, the WACC of the classifier built from a watermarked pre-trained encoder should be always higher than $\mathcal{T}$.
%    The clean accuracy is the accuracy of a downstream classifier based on a clean encoder.
%    \item \textbf{Watermarked Accuracy (WA)}: The watermarked accuracy is the accuracy of a downstream classifier based on a watermarked encoder.    \item \textbf{Watermark Recognizing Rate (WRR)}: According to our method, the 'right' trigger will cause the watermarked encoder to produce anomalous feature vectors. Therefore, some images which can be correctly classified at first will be misclassified after being embedded with the trigger. The WRR is the ratio of these wrongly classified images to the ones which are rightly classified if there's no trigger embedded.
\end{itemize}

%We use {\it Clean Accuracy (CA)}, {\it Watermarked Accuracy (WA)}, {\it Watermark Recognizing Rate (WRR)} to evaluate our watermarked model. {\it Clean Accuracy} represents the proportion of correctly-classified images by a pre-trained model without watermark embedding. {\it Watermarked Accuracy} represents the proportion of correctly-classified images by a classifier based on a watermarked pre-trained encoder. {\it Watermark Recognizing Rate} shows the effectiveness of our method by testing encoder-based classifiers using trigger embedded images $x_i \oplus e_t$. The three coefficients are defined as below:

\subsection{Uniqueness Analysis}

To show the uniqueness of our watermark, we carry out our experiment by evaluating the WACC for two aspects. 
First, we use a set of triggered samples to query the clean model and the corresponding downstream classifier built from the watermarked pre-trained encoder respectively. 
Second, we use a set of randomly designed wrong triggered samples to test a watermarked downstream classifier.
The evaluation results are shown in Table \ref{table:1} and Table~\ref{table:2}. 

\begin{table}[!htbp]
\centering
\resizebox{0.9\linewidth}{!}{
    \begin{tabular}{c|c|c|c} \Xhline{1pt}
    Pre-train   &
    Downstream   &
    \multicolumn{2}{c}{\textbf{Model WACC (\%)}} 
    \\\cline{3-4}
    method& dataset& Clean model & Watermarked model\\ \Xhline{1pt}
    
    & STL10 & 22.09 & 91.76 \\ 
    SimCLR & GTSRB & 37.43 & 93.37\\
    & SVHN  & 54.94 & 80.13\\ \hline
    
    & STL10 & 7.01 & 90.51 \\ 
    MoCoV2 & GTSRB & 12.47 & 93.92\\
    & SVHN  & 4.52 & 84.37\\ \hline
    
    \Xhline{1pt}
    \end{tabular}}
    \vspace{-1ex}
    \caption{WACC of clean models and watermarked models.}
    \label{table:1}
\end{table}

\begin{table}[!htbp]
\centering
\resizebox{0.9\linewidth}{!}{
    \begin{tabular}{c|c|c|c} \Xhline{1pt}
    Pre-train   &
    Downstream   &
    \multicolumn{2}{c}{\textbf{Model WACC (\%)}} 
    \\\cline{3-4}
    method& dataset& Wrong trigger & Correct trigger \\ \Xhline{1pt}
    
    & STL10 & 8.34 & 91.76 \\ 
    SimCLR & GTSRB & 22.32 & 93.37\\
    & SVHN  & 41.90 & 81.90\\ \hline
    
    & STL10 & 8.17 & 90.51 \\ 
    MoCoV2 & GTSRB & 9.33 & 93.92\\
    & SVHN  & 2.17 & 84.37\\ \hline
    
    \Xhline{1pt}
    \end{tabular}}
    \vspace{-1ex}
    \caption{WACC of different triggers on the watermarked models.}
    \label{table:2}
\end{table}

The results indicate that our method can embed an efficient watermark into the clean pre-trained encoder and achieve high uniqueness. 
As shown in Table~\ref{table:1}, our watermarking method can effectively distinguish the clean and watermarked model by giving significantly different WACC for triggered samples. 
Also, the difference between the AC of the `correct' trigger and the AC of the `wrong' trigger is also prominent. 
This means for wrong triggered samples, our watermarked model will give a very low WACC for a successful IP verification with uniqueness guaranteed.

%On the one hand, there is a distinct difference between WA of clean models and WA of watermarked models when both models are tested by images embedded with a corresponding trigger. On the other hand, the difference between the AC of the `correct' trigger and the AC of the `wrong' trigger is also prominent. In other words, a watermarked model can only recognize the corresponding trigger.

\subsection{Robustness analysis} 
To test the Robustness of our watermark, we calculate ACC and WACC of our watermarked model which is processed with fine-tuning or pruning by the attackers. 
The results are shown in Table \ref{table:3} and Table \ref{table:4}. 

\begin{table}[!htbp]
\centering
\resizebox{0.65\linewidth}{!}{
    \begin{tabular}{c|c|c} \Xhline{1pt}
    \makecell{Pruning ratio}   & ACC (\%) & \makecell{WACC (\%)} \\ \Xhline{1pt}
    
    \makecell[c]{0.2} & \makecell[c]{76.93} & \makecell[c]{89.43}\\
    \makecell[c]{0.4} & \makecell[c]{76.15} & \makecell[c]{88.33}\\
    \makecell[c]{0.6} & \makecell[c]{73.68} & \makecell[c]{88.06}\\
    \makecell[c]{0.8} & \makecell[c]{64.96} & \makecell[c]{59.95}\\
    \makecell[c]{0.9} & \makecell[c]{55.75} & \makecell[c]{43.75}\\
    \makecell[c]{0.95} & \makecell[c]{46.39} & \makecell[c]{28.45}\\ \hline
    
    \Xhline{1pt}
    \end{tabular}}
    \vspace{-1ex}
    \caption{Robustness evaluation against model pruning on SimCLR.}
    \label{table:3}
\end{table}

\begin{table}[!htbp]
\centering
\resizebox{0.8\linewidth}{!}{
    \begin{tabular}{c|c|c|c} \Xhline{1pt}
    \makecell{Fine-tune \\ method}&\makecell{Downstream \\ dataset}& ACC (\%) & \makecell{ WACC (\%)} \\ \Xhline{1pt}
    
    & STL10 & 80.74 & 83.15 \\ 
    FTAL & GTSRB & 63.15 & 81.87\\
    & SVHN  & 94.34 & 87.04\\ \hline
    
    & STL10 & 76.27 & 91.76 \\ 
    RTLL & GTSRB & 82.43 & 93.92\\
    & SVHN  & 66.82 & 84.37\\ \hline
    
    \Xhline{1pt}
    \end{tabular}}
    \vspace{-1ex}
    \caption{Robustness evaluation against fine-tuning.}
    \label{table:4}
\end{table}

From Table \ref{table:3} we can observe that the WACC keeps at a high level with the pruning amount increasing when the pruning process still has a slight impact on ACC. 
When the pruning ratio is beyond 0.8, there is an immediate drop in both ACC and WACC of the watermarked models. 
This indicates that a very large pruning ratio can mitigate the watermarking method but also significantly compromise the model's functionality.
Such a high pruning ratio is pointless for an attacker since a broken model is useless. 
Thus, we conclude that our watermarking method is robust against model pruning.

Table \ref{table:4} shows the ACC and WACC of the models fine-tuned by RTLL and FTAL. 
The WACC has only a slight drop when the model is fine-tuned which indicates that our watermarking method is robust against model fine-tuning. 
%It can be seen that the WACC does drop slightly when the model is fine-tuned with FTAL, but are still beyond 80\%. Thus, we claim strong robustness of our watermark.

\subsection{Performance-preserving analysis}
We measure the performance-preserving requirement by comparing the ACC of the clean model and the watermarked one. The results are shown in Table \ref{table:5}. 

\begin{table}[!htbp]
\centering
\resizebox{0.8\linewidth}{!}{
    \begin{tabular}{c|c|c|c} \Xhline{1pt}
    Pre-train   &
    Downstream   &
    \multicolumn{2}{c}{\textbf{Model ACC (\%)}} 
    \\\cline{3-4}
    method& Dataset& Clean & Watermarked \\ \Xhline{1pt}
    
    & STL10 & 76.14 & 76.11 \\ 
    SimCLR & GTSRB & 81.40 & 82.43\\
    & SVHN  & 66.82 & 66.20\\ \hline
    
    & STL10 & 89.16 & 90.68 \\ 
    MoCoV2 & GTSRB & 75.96 & 76.12\\
    & SVHN  & 79.41 & 77.31\\ \hline
    
    \Xhline{1pt}
    \end{tabular}}
    \vspace{-1ex}
    \caption{Functionality evaluation results by comparing the ACC of clean model with the watermarked model.}
    \label{table:5}
\end{table}
\vspace{-1ex}

It turns out that our watermark embedding algorithm preserves the functionality of the pre-trained encoder. 
We can discover for Table \ref{table:5} that the ACC of watermarked models is similar or even larger than that of clean models in most cases. 
We analyze that an even larger ACC with watermarking is due to the phenomenon that training on noisy data gives significant robustness improvements pointed in~\cite{radford2021learning}. 
The noise brought by the triggered samples in the training dataset will not harm the model ACC which proves the robustness of our method. 

\section{Related Works}

Numerous watermarking schemes have been proposed for conventional DNN models.
They can be roughly classified into the following two categories:

%\vspace{3pt}
\noindent\textbf{White-box solutions.} 
This strategy adopts redundant bits as watermarks and embeds them into the model parameters. For instance, \cite{uchida2017embedding} introduced a parameter regularizer to embed a bit-vector (e.g. signature) into model parameters which can guarantee the performance of the watermarked model. \cite{rouhani2019deepsigns} found that implanting watermarks into model parameters directly could affect their static properties (e.g histogram). Thus, they injected watermarks in the probability density function of the activation sets of the DNN layers.  These methods require the owner to have white-box access to the model parameters during the watermark extraction and verification phase, which can significantly limit the possible usage scenarios.

\vspace{3pt}
\noindent\textbf{Black-box solutions.} 
This strategy takes a set of unique sample-label pairs as watermarks and embeds their correlation into DNN models. For example, \cite{le2019adversarial} adopted adversarial examples near the frontiers as watermarks to identify the ownership of DNN models. Zhang et al. \cite{zhang2018protecting} and Adi et al. \cite{adi2018turning} employed backdoor attack techniques to embed backdoor samples with certain trigger patterns into DNN models. \cite{namba2019robust} and \cite{li2019prove} generated watermark samples that are almost indistinguishable from normal samples to avoid detection by adversaries. \cite{ChenGuo21} designed temporal state sequences to watermark reinforcement learning models. \cite{xiaoxuan2021meets} utilized cache side channels to verify watermarks in the model architecture. 

In addition to watermarking, an alternative approach to IP protection is model fingerprinting. The main difference is that the model owner crafts special samples to verify the model without making any modifications to it. 

\vspace{3pt}
\noindent\textbf{DNN Fingerprint.}
Some works proposed fingerprinting schemes for classification models. They leveraged adversarial examples as fingerprints to identify the target models. For example, IPGuard~\cite{cao2019ipguard} identified the data samples close to the target model's decision boundary to fingerprint this model.~\cite{lukas2019deep} adopted conferrable adversarial examples with certain labels, which can transfer from the target model to its surrogates while remaining ineffective to other non-target models. \cite{he2019sensitive} designed sensitive samples to fingerprint black-box models, which will fail even the model has very small modifications. 

\vspace{3pt}
\section{Conclusion and Future Work}

In this paper, we propose a novel watermarking technique to protect the IP of pre-trained encoders via contrastive learning. We introduce a new loss function, which can effectively embed the watermark into the encoder without the knowledge of the downstream tasks and datasets. The watermark can be transferred to any downstream model built from this encoder. We perform extensive evaluations to demonstrate our watermarking methodology has high uniqueness, functionality-preserving, and robustness. 

We have the following directions as future work. (1) We aim to develop an algorithm to optimize our trigger pattern, to reduce the false positives of the watermark verification. (2) We can co-optimize the trigger and encoder to further enhance the uniqueness of the watermark. (3) We will design more effective attack methods to detect or remove any watermarks in a pre-trained encoder. 

%We can 1) develop an algorithm to optimize our trigger so that it will have less influence on WA of the clean model, 2) develop a trigger generating algorithm to enhance the uniqueness, and 3)develop an algorithm to wipe out the watermark so as to disable the verification samples.

\clearpage
\bibliographystyle{named}
\bibliography{ijcai19}

\end{document}